\documentclass[10pt,twocolumn,letterpaper]{article}

\usepackage{cvpr}
\usepackage{times}
\usepackage{epsfig}
\usepackage{graphicx}
\usepackage{amsmath}
\usepackage{amssymb}

\usepackage{authblk}
\usepackage{adjustbox}
\usepackage{color}
\usepackage{xcolor}
\usepackage{url}
\usepackage{wrapfig}

\usepackage{multirow}

\newcommand{\widthscalefive}{0.125}
\usepackage{booktabs}

\usepackage{soul}

\usepackage[pagebackref=true,breaklinks=true,letterpaper=true,colorlinks,bookmarks=false]{hyperref}

\cvprfinalcopy 

\def\cvprPaperID{4731} 

\ifcvprfinal\fi
\begin{document}

\title{TDAN: Temporally Deformable Alignment Network for Video Super-Resolution}

\author[1]{Yapeng Tian}
\author[2]{Yulun Zhang}
\author[2,3]{Yun Fu}
\author[1]{Chenliang Xu}
\affil[1]{Department of Computer Science, University of Rochester}
\affil[2]{Department of Electrical and Computer Engineering, Northeastern University}
\affil[3]{College of Computer and Information Science, Northeastern University}
\maketitle

\begin{abstract}
Video super-resolution (VSR) aims to restore a photo-realistic high-resolution (HR) video frame from both its corresponding low-resolution (LR) frame (reference frame) and multiple neighboring frames (supporting frames). Due to varying motion of cameras or objects, the reference frame and each support frame are not aligned. Therefore, temporal alignment is a challenging yet important problem for VSR. Previous VSR methods usually utilize optical flow between the reference frame and each supporting frame to wrap the supporting frame for temporal alignment. Therefore, the performance of these image-level wrapping-based models will highly depend on the prediction accuracy of optical flow, and inaccurate optical flow will lead to artifacts in the wrapped supporting frames, which also will be propagated into the reconstructed HR video frame. To overcome the limitation, in this paper, we propose a temporal deformable alignment network (TDAN) to adaptively align the reference frame and each supporting frame at the feature level without computing optical flow. The TDAN uses features from both the reference frame and each supporting frame to dynamically predict offsets of sampling convolution kernels. By using the corresponding kernels, TDAN transforms supporting frames to align with the reference frame. To predict the HR video frame, a reconstruction network taking aligned frames and the reference frame is utilized.  Experimental results demonstrate the effectiveness of the proposed TDAN-based VSR model.
\end{abstract}

\section{Introduction}

\begin{figure}
	\scriptsize
	\centering
    
    \begin{adjustbox}{valign=t}
			\begin{tabular}{c}
				\includegraphics[width=0.204\textwidth]{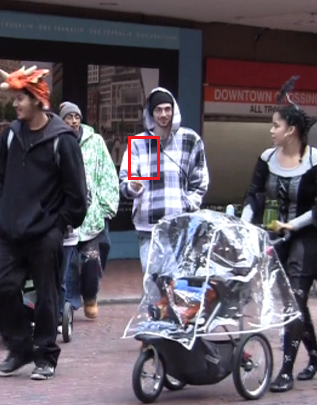}
				\\
				\textit{Walk}
			\end{tabular}
	\end{adjustbox}
	\hspace{-2.3mm}
		\begin{adjustbox}{valign=t}
			\begin{tabular}{cccc}
				\includegraphics[width=0.09\textwidth]{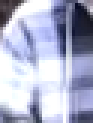} \hspace{-1.5mm} &
				\includegraphics[width=0.09\textwidth]{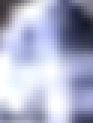}
				\\
				HR \hspace{-1.5mm} &
				Bicubic
				\\
		
				\includegraphics[width=0.09\textwidth]{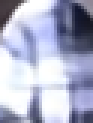} \hspace{-1.5mm} &
				\includegraphics[width=0.09\textwidth]{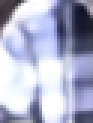}
				\\
				DUF~\cite{jo2018deep} \hspace{-1.5mm} &
				Ours
				\\
			\end{tabular}
			\end{adjustbox}

	\caption{VSR results for a frame in the \textit{walk} sequence. We can find that our method can restore more accurate image structures and details than the recent DUF network.}
	\label{fig:democase}
\end{figure}

The goal of video super-resolution (VSR) is to reconstruct a high-resolution (HR) video frame from its corresponding low-resolution (LR) video frame (reference frame) and multiple neighboring LR video frames (supporting frames). HR video frames contain more image details and are more preferred to humans. Therefore, the VSR technique is desired in many real applications, such as video surveillance and high-definition television (HDTV).

To super-resolve the LR reference frame, VSR will exploit both the LR reference frame and multiple LR supporting frames. However, the LR reference frame and each supporting frame are likely not fully aligned due to the motion of camera or objects. Thus, a vital issue in VSR is how to align the supporting frames with the reference frame.

Previous methods~\cite{caballero2017real,liu2017robust,tao2017detail,xue2017video,sajjadi2018frame} usually exploit optical flow to predict motion fields between the reference frame and supporting frames, then wrap the supporting frames using their corresponding motion fields. Therefore, the optical flow prediction is crucial for these approaches, and any errors in the flow computation or the image-level wrapping operation may introduce artifacts around image structures in the aligned supporting frames.

To alleviate the above issues, we propose a temporally deformable alignment network (TDAN) in this paper that performs one-stage temporal alignment without using optical flow. Unlike previous optical flow-based VSR methods, our approach can adaptively align the reference frame and supporting frames at a feature level without explicit motion estimation and image wrapping operations. Therefore, the aligned LR video frames will have less annoying image artifacts, and the image quality of the reconstructed HR video frame will be improved. In specific, inspired by the deformable convolution~\cite{Dai_2017_ICCV}, the proposed TDAN uses features from both the reference frame and each supporting frame to dynamically predict offsets of sampling convolution kernels, and then apply these dynamic kernels on features from supporting frames to employ the temporal alignment. Here, given different reference and supporting frame pairs, the module will generate their corresponding sampling kernels, which makes the TDAN have strong capability and flexibility to handle various motion conditions in temporal scenes. With the aligned supporting frames and the reference frame, a reconstruction network is utilized to predict an HR video frame corresponding to the LR reference frame.

We conduct extensive experiments on a widely-used VSR benchmark and two real-world LR video sequences. The experimental results show that our framework achieves state-of-the-art performance. In Fig.~\ref{fig:democase}, we show a visual comparison to the recent DUF network \cite{jo2018deep}, and we observe that our method reconstructs more image details.

The contributions of this paper are three-fold: (1) we propose a novel temporally deformable alignment network (TDAN) for feature-level alignment, which avoids the two-stage process adopted by previous optical flow-based methods;
(2) we propose an end-to-end trainable VSR framework based on the TDAN; and (3) our method achieves state-of-the-art VSR performance on Vid4 \cite{liu2014bayesian} benchmark dataset. The source code and pre-trained models will be released. A video demo is available on \textcolor{magenta}{\url{https://www.youtube.com/watch?v=eZExENE50I0}}.

\section{Related Work}

In this section, we discuss three related works: single image super-resolution (SISR), video super-resolution (VSR), and deformable convolution~\cite{Dai_2017_ICCV}.

\vspace{2mm}
\noindent \textbf{Single Image Super-Resolution (SISR):}
Dong \etal~\cite{dong2014learning} firstly proposed an end-to-end image super-resolution convolutional neural network (SRCNN). Kim \etal~\cite{kim2016accurate} introduced a 20-layer deep network: VDSR with residual learning. Shi \etal~\cite{shi2016real} learned an efficient sub-pixel convolution layer to upscale the final LR feature maps into the HR output for accelerating SR networks. Deeper networks like LapSRN~\cite{lai2017deep}, DRRN~\cite{tai2017image}, and MemNet~\cite{tai2017memnet}, were explored to further improve SISR performance. However, training images used in the previous methods have limited resolution, which makes the training of even deeper and wider networks very difficult. Recently, Timofte \etal introduced a novel large dataset (DIV2K) consisting of 1000 DIVerse 2K resolution RGB images in NTIRE 2017 Challenge~\cite{timofte2017ntire}. Current state-of-the-art SISR networks, like EDSR~\cite{lim2017enhanced}, RDN~\cite{zhang2018residual}, DBPN~\cite{haris2018deep}, and RCAN~\cite{zhang2018image}, trained on the DIV2K outperformed previous networks by a substantial margin. A recent survey about deep learning-based SISR methods is in~\cite{yang2018deep}.

\vspace{2mm}
\noindent \textbf{Video Super-Resolution (VSR):}
It has been observed that temporal alignment critically affects the performance of VSR systems. Previous methods usually adopted two-stage approaches based on optical flow. They will conduct motion estimation by computing optical flow in the first stage and utilize the estimated motion fields to perform image wrapping/motion compensation in the second stage. For example, Liao \etal~\cite{Liao_2015_ICCV} used two classical optical flow methods: TV-$L_1$ and MDP flow~\cite{xu2012motion} with different parameters to generate HR SR-drafts, and then predicted the final HR frame by a deep draft-ensemble network. Kappeler \etal~\cite{kappeler2016video} took interpolated flow-wrapped frames as inputs of a CNN to predict HR video frames. However, the pioneering methods both used classical optical flow algorithms, which are separated from the frame reconstruction CNN and much slower than the flow CNN during inference.

To address the issue, Caballero \etal~\cite{caballero2017real} introduced the first end-to-end VSR network:~VESCPN, which jointly trains flow estimation and spatiotemporal networks. Liu \etal~\cite{liu2017robust} proposed a temporal adaptive neural network to adaptively select the optimal range of temporal dependency and a rectified optical flow alignment method for better motion estimation. Tao \etal~\cite{tao2017detail} computed LR motion field based on optical flow network and designed a new layer to utilize sub-pixel information from motion and simultaneously achieve sub-pixel motion compensation (SPMC) and resolution enhancement. Xue \etal~\cite{xue2017video} exploited task-oriented motion cues via Task-Oriented Flow (TOFlow), which achieves better VSR results than fixed flow algorithms. Sajjadi \etal~\cite{sajjadi2018frame} extended the conventional VSR models to a frame-recurrent VSR framework.
%
However, sufficiently high-quality motion estimation is not easy to obtain even with state-of-the-art optical flow estimation networks. Even with accurate motion fields, the image-wrapping based motion compensation will produce artifacts around image structures, which may be propagated into final reconstructed HR frames. The proposed TDAN performs a feature-wise one-stage temporal alignment without relying on optical flow, which will alleviate the issues in these previous optical flow-based VSR networks.

\begin{figure*}
  \centering
    \includegraphics[width=\textwidth]{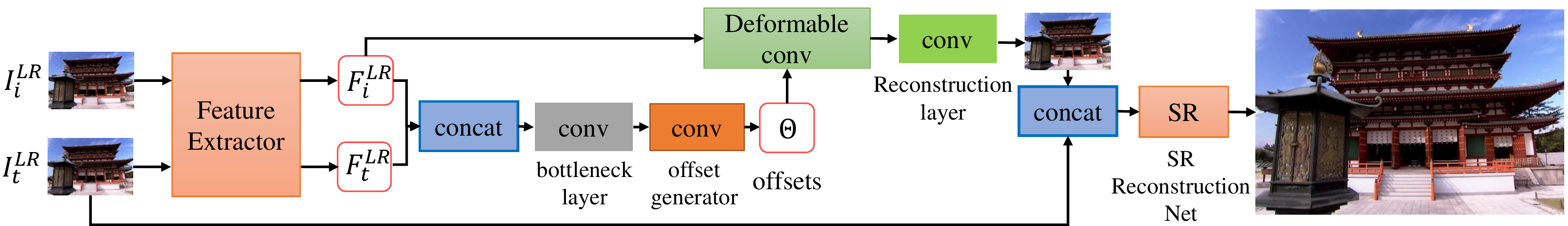}
    \caption{The proposed TDAN-based VSR framework. Here, we only show the framework with one supporting frame. In our implementation, 4 neighboring supporting frames are exploited for exploring more temporal information.}
\label{framework}
\end{figure*}

\vspace{2mm}
\noindent \textbf{Deformable Convolution:}
CNNs have the inherent limitation in modeling geometric transformations due to the fixed kernel configuration. For enhancing the transformation modeling capability of regular CNNs, Dai \etal~\cite{Dai_2017_ICCV} proposed a deformable convolution operation. It has been applied to address several high-level vision tasks such as object detection~\cite{Dai_2017_ICCV}, video object detection~\cite{Bertasius_2018_ECCV}, semantic segmentation~\cite{Dai_2017_ICCV}, and human pose estimation~\cite{Sun_2018_ECCV}.
Although the deformable convolution has shown superiority on several high-level vision tasks, it was rarely explored for low-level vision problems.

\section{Method}

First, we define the problem and give an overview of our framework in Sec.~\ref{overview}. Upon this framework, we propose the temporally deformable alignment network in Sec.~\ref{TDAN}, the SR reconstruction network in Sec.~\ref{recon}, and the loss functions in Sec.~\ref{loss}. Finally, in Sec.~\ref{analysis}, we discuss the merits of the proposed alignment network.

\subsection{Overview}
\label{overview}

Let $I_{t}^{LR}\in\mathbb{R}^{H\times W\times C }$ be the $t$-th LR video frame, and $I_{t}^{HR}\in\mathbb{R}^{sH\times sW\times C}$ be the corresponding HR video frame, where $s$ is the upscaling factor. Our goal is to restore the HR video frame $I_{t}^{HR}$ from the reference LR frame $I_{t}^{LR}$ and $2N$ supporting LR frames $\{I_{t-N}^{LR},... I_{t-1}^{LR}, I_{t+1}^{LR},...,I_{t+N}^{LR}\}$. Therefore, our VSR framework takes the consecutive $2N+1$ frames $\{I_{i}\}_{i=t-N}^{t+N}$ as the input to predict the HR frame $I_{t}^{HR}$, which is illustrated in Fig.~\ref{framework}. It consists of two main sub-networks: a temporally deformable alignment network (TDAN) to align each supporting frame with the reference frame and a SR reconstruction network to predict the HR frame.

The TDAN takes an LR supporting frame $I_{i}^{LR}$ and the LR reference frame $I_{t}^{LR}$ as inputs to predict the corresponding aligned LR frame $I_{i}^{LR'}$ for the supporting frame:
\begin{equation}
I_{i}^{LR'} = f_{TDAN}(I_{t}^{LR}, I_{i}^{LR})
\enspace.
\label{tdan_eq}
\end{equation}
Feeding the $2N$ supporting frames into the TDAN separately, we can obtain $2N$ corresponding aligned LR frames $\{I_{t-N}^{LR'},... I_{t-1}^{LR'}, I_{t+1}^{LR'},...,I_{t+N}^{LR'}\}$.

The SR reconstruction network will utilize the $2N$ aligned frames along with the reference frame to restore the HR video frame $I_{t}^{HR} = f_{SR}(I_{t-N}^{LR'},..., I_{t-1}^{LR'},I_{t}^{LR}, I_{t+1}^{LR'},...,I_{t+N}^{LR'})$.

\subsection{Temporally Deformable Alignment Network}
\label{TDAN}
Given an LR supporting frame $I_{i}^{LR}$ and the LR reference frame $I_{t}^{LR}$, the proposed TDAN will temporally align $I_{i}^{LR}$ with $I_{t}^{LR}$. It mainly consists of three modules: feature extraction, deformable alignment, and aligned frame reconstruction.

\noindent \textbf{Feature Extraction:} This module extracts visual features $F_{i}^{LR}$ and $F_{t}^{LR}$ from $I_{i}^{LR}$ and $I_{t}^{LR}$ respectively via a shared feature extraction network. The network consists of one convolutional layer and $k_1$ residual blocks~\cite{he2016deep} with ReLUs as the activation functions. In our implementation, we adopted a modified residual structure from~\cite{lim2017enhanced}. The extracted features will be utilized for feature-wise temporal alignment.

\noindent \textbf{Deformable Alignment:} The deformable alignment module takes the $F_{i}^{LR}$ and $F_{t}^{LR}$ as inputs to predict sampling parameters $\Theta$ for the feature $F_{i}^{LR}$:
\begin{equation}
\Theta = f_{\theta}(F_{i}^{LR}, F_{t}^{LR})
\enspace.
\label{theta_eq}
\end{equation}
Here, $\Theta=\{\triangle p_n \mid n=1,\dots,|\mathcal{R}|\}$ refers to the offsets of the convolution kernels, where $\mathcal{R}=\{(-1, -1),(-1,0),...,(0, 1),(1, 1)\}$ donates a regular grid of a $3\times3$ kernel. With $\Theta$ and $F_{i}^{LR}$, the aligned feature $F_{i}^{LR'}$ of the supporting frame can be computed by the deformable convolution:
\begin{equation}
F_{i}^{LR'}=f_{dc}(F_{i}^{LR}, \Theta)
\enspace.
\label{theta_eq}
\end{equation}
More specifically, for each position $p_0$ on the aligned feature map $F_{i}^{LR'}$, we have:
\begin{equation}
F_{i}^{LR'}(p_0)=\sum_{p_n\in\mathcal{R}}w(p_n)F_{i}^{LR}(p_0+p_n+\triangle p_n)
\enspace.
\label{aligned_fea_eq}
\end{equation}
The convolution will be operated on the irregular positions $p_n+\triangle p_n$, where the $\triangle p_n$ may be fractional. To address the issue, the operation is implemented by using bilinear interpolation, which is the same as that proposed in \cite{Dai_2017_ICCV}.

Here, the deformable alignment module consists of several regular and deformable convolutional layers. For the sampling parameter generation function $f_{\theta}$, it concatenates $F_{i}^{LR}$ and $ F_{t}^{LR}$, and uses a $3\times3$ bottleneck layer to reduce the channel number of the concatenated feature map. Then, the sampling parameters are predicted by a convolutional layer with the kernel size $|\mathcal{R}|$ as the output channel number. Finally, the aligned feature $F_{i}^{LR'}$ is obtained from $\Theta$ and $F_{i}^{LR}$ based on deformable convolution operation. In practice, we use three additional deformable convolutional layers before and after the $f_{dc}$ for enhancing the transformation flexibility and capability of the module. Section~\ref{ablation} contains the ablation study on the performance of the module with different numbers of the additional deformable convolutional layers.

We note that the feature of the reference frame $F_{t}^{LR}$ is only used for computing the offset, and its information will not be propagated into the aligned feature of the supporting frame $F_{i}^{LR'}$. Besides, the adaptively learned offset will implicitly capture motion cues and also explore neighboring features within the same image structures for temporal alignment.

\noindent \textbf{Aligned Frame Reconstruction:}  Although the deformable alignment has the potential to capture motion cues and align $F_{i}^{LR}$ with $F_{t}^{LR}$, the implicit alignment is very difficult to learn without a supervision. To make the implicit alignment be possible, we restore an aligned LR frame $I_{i}^{LR'}$ for $I_{i}^{LR}$ and utilize a alignment loss for enforcing the deformable alignment module to capture motions and align two frames at the feature level. More discussions can be found in Sec.~\ref{lim_fail}. The aligned LR frame $I_{i}^{LR'}\in\mathbb{R}^{H\times W\times C }$ can be reconstructed from the aligned feature map with a $3\times3$ convolutional layer.

After feeding $2N$ reference and supporting frame pairs into the TDAN, we can obtain the corresponding $2N$ aligned LR frames, which will be used to predict the HR video frame $I_{t}^{HR}$ in the SR reconstruction network.

\subsection{SR Reconstruction Network}
\label{recon}

We use a SR reconstruction network to restore the HR video frame  $I_{i}^{HR}$ from the aligned LR frames and the reference frame.
The network contains three modules: temporal fusion, nonlinear mapping, and HR frame reconstruction, which will aggregate temporal information from different frames, predict high-level visual features, and restore the HR frame for the LR reference frame, respectively.

\noindent \textbf{Temporal Fusion:} To fuse different frames cross the time-space, we directly concatenate the $2N+1$ frames and then feed them into a $3\times3$ convolutional layer to output the fused feature map.

\noindent \textbf{Nonlinear Mapping:} The nonlinear mapping module with $k_2$ stacked residual blocks~\cite{lim2017enhanced}  will take the shadow fused features as input to predict deep features.

\noindent \textbf{HR Frame Reconstruction:} After extracting deep features in the LR space, inspired by the EDSR~\cite{lim2017enhanced},  we utilize an upscaling layer to increase the resolution of the feature map with a sub-pixel convolution as proposed by Shi \etal~\cite{shi2016real}. In practice, for $\times4$ upscaling, two sub-pixel convolution modules will be used. The final HR video frame estimation $I_{t}^{HR'}$ will be obtained by a convolutional layer from the zoomed feature map.

\subsection{Loss Function}
\label{loss}

Two loss functions $\mathcal{L}_{align}$ and $\mathcal{L}_{sr}$ are used to train the TDAN and SR reconstruction network, respectively. Note that we have no groundtruth of the aligned LR frames. To optimize the TDAN, we utilize the reference frame as the label and make the aligned LR frames close to the reference frame:
\begin{equation}
\mathcal{L}_{align}=\frac{1}{2N}\sum_{i=t-N, \neq t}^{t+N}\parallel I_{i}^{LR'} -I_{t}^{LR} \parallel_{1}
\enspace.
\label{align_loss}
\end{equation}
The objective function of the SR reconstruction network is defined via $\mathcal{L}_1$ reconstruction loss:
\begin{equation}
\mathcal{L}_{sr}=\parallel I_{t}^{HR'} -I_{t}^{HR}\parallel_{1}
\enspace.
\label{sr_loss}
\end{equation}
Combining the two loss terms, we have the overall loss function for training our VSR framework:
\begin{equation}
\mathcal{L}=\mathcal{L}_{align} + \mathcal{L}_{sr}
\enspace.
\label{overall_loss}
\end{equation}
The two loss terms are simultaneously optimized when training our VSR framework. Therefore, our TDAN-based VSR network is end-to-end trainable. In addition, the TDAN can be trained with a self-supervision without requiring any annotations.

\subsection{Analyses of the Proposed TDAN}
\label{analysis}

Given a reference frame and a set of supporting frames, the proposed TDAN can employ temporal alignment to align the supporting frames with the reference frame. It has several merits:

\noindent \textbf{One-Stage Temporal Alignment:} Most previous temporal alignment methods are based on optical flow, which will split the temporal alignment problem into two sub-problems: flow/motion estimation and motion compensation. As discussed in the paper, the performance of these methods will highly rely on the accuracy of flow estimation, and the flow-based image wrapping will introduce annoying artifacts. Unlike these two-stage temporal alignments, our TDAN is a one-stage approach, which aligns the supporting frames at the feature level. It implicitly captures motion cues via adaptive sampling parameter generation without explicitly estimating the motion field, and reconstructs the aligned frames from the aligned features.

\noindent \textbf{Self-Supervised Training:} The optical flow estimation is crucial for the two-stage methods. For ensuring the accuracy of flow estimation, some VSR networks~\cite{tao2017detail,xue2017video,liu2017robust} utilized additional flow estimation algorithms. Unlike these methods, there is no flow estimation inside the TDAN, and it can be trained in a self-supervised manner without relying on any extra supervisions.

\noindent \textbf{Exploration:} For each location in a frame, its motion field computed by optical flow only refers to one potential position $p$. It means that each pixel in a wrapped frame will only copy one pixel at $p$ or use an interpolated value for a fractional position. However, beyond utilizing information at $p$, our deformable alignment module can adaptively explore more features at sampled positions, which may share the same image structure as $p$, and it will help to aggregate more contexts for better-aligned frame reconstruction. Therefore, the proposed TDAN has stronger exploration capability than optical flow-based models.

\noindent \textbf{Generic:} The proposed TDAN is a generic temporal alignment framework and can be easily utilized to replace flow-based motion compensation for several other tasks, such as video denoising~\cite{mahmoudi2005fast}, video deblocking~\cite{maggioni2012video}, video deblurring~\cite{hyun2015generalized}, video frame interpolation~\cite{ascenso2005improving}, and even video prediction~\cite{wichers2018hierarchical}.

\section{Experiments}
\label{exp}
First, we introduce the experimental settings: datasets, evaluation metrics, degradation methods, and training settings in Sec.~\ref{exp_setting}. Then, we compare the proposed VSR framework with other state-of-the-art VSR and SISR approaches on different degradation for fair comparisons in Sec.~\ref{exp_comp}. Furthermore, ablation studies on different settings of our TDAN are shown in Sec.~\ref{ablation}. Finally, some results on real-world videos are illustrated in Sec.~\ref{real}.

\subsection{Experimental Settings}
\label{exp_setting}
\noindent \textbf{Datasets:}
A large-scale dataset is important for training high-performance networks. For SR tasks, besides the scale of the training data, resolution of the training images is another important factor. With the emergence of the DIV2K dataset~\cite{timofte2017ntire}, which contains 1000 2K-resolution HR images, training very deep network (\eg beyond 400 convolutional layers in RCAN~\cite{zhang2018image}) or very wide network (\eg EDSR~\cite{lim2017enhanced}) becomes possible. As a result, the SISR performance has been greatly improved recently. Similar to SISR, for training a high-capability VSR network, a large video dataset containing very HR frames is desired. Previous methods~\cite{liu2017robust,tao2017detail,caballero2017real,sajjadi2018frame,jo2018deep} usually collect 720P, 1080P, or even 4K videos from the internet, and then crop training patches from the collected samples. Jo \etal~\cite{jo2018deep} created a large dataset with various contents including wildlife, activity, and landscape, and augmented the dataset by a temporal augmentation method. Sajjadi \etal introduced another high-quality VSR dataset in~\cite{sajjadi2018frame}. However, these datasets are all not publicly available\footnote{Although Sajjadi \etal provided the URLs of used videos, some videos are not accessible, and the raw data needs additional pre-processing and cleaning.}. Thanks to Xue~\etal~\cite{xue2017video} for releasing a VSR dataset: Vimeo Super-Resolution dataset containing 64612 training samples. Each sample in the dataset contains seven consecutive frames with $448 \times 256$ resolution. However, unlike the previous methods~\cite{sajjadi2018frame,jo2018deep}, the authors did not create HR training samples by cropping full-resolution frames (or slightly down-sampled frames). Instead, they resized the original frames to $448 \times 256$ resolution, which will smooth out many image details.

In our experiments, we used Vimeo Super-Resolution dataset as our training dataset and 31 frames from the Temple sequence~\cite{Liao_2015_ICCV} as the validation dataset. We evaluated our models on the Vid4 benchmark~\cite{liu2014bayesian}, which contains four video sequences: \textit{city}, \textit{walk}, \textit{calendar}, and \textit{foliage}, as other methods.

\noindent \textbf{Evaluation Metrics:}
PSNR and SSIM~\cite{wang2004image} are used as evaluation metrics to compare different VSR networks quantitatively. Excepting comparison with the TOFlow, the first and last two frames are not used for evaluation, and four spatial boundary pixels are ignored.

\noindent \textbf{Degradation Methods:}
We compare our TDAN-based network with current state-of-the-art VSR and SISR networks: VSRnet~\cite{kappeler2016video}, ESPCN~\cite{shi2016real}, VESCPN~\cite{caballero2017real}, Liu~\etal~\cite{liu2017robust}, TOFlow~\cite{xue2017video}, DBPN~\cite{haris2018deep}, RND~\cite{zhang2018residual}, RCAN~\cite{zhang2018image}, SPMC~\cite{tao2017detail}, FSRVSR~\cite{sajjadi2018frame}, and DUF~\cite{jo2018deep}. The first seven networks adopted the Matlab function \textit{imresize} with the option bicubic (BI) to generate LR video frames. SPMC, FSRVSR, and DUF obtained LR frames by first blurring HR frames via a Gaussian kernel and
then downsampling via selecting every $s$-th pixel (denote as BD for short). Note that we compare the FRVSR-3-64 and DUF-16L models, which have similar model sizes as our TDAN-based VSR network. We train two different TDAN models with the two different degradation methods for fair comparisons.

\noindent \textbf{Training Settings:}
We use a downsampling factor: $s$ = 4 in all our experiments. Each training batch contains $64\times5$ LR RGB patches with the size $48\times48$, where 64 means the batch size and $5$ refers to the number of consecutive input frames. We implement our network with Pytorch~\cite{paszke2017automatic} and adopt Adam~\cite{kingma2014adam} as the optimizer. The learning rate is initialized to $10^{-4}$ for all layers and decreases half for every 100 epochs. For every 100 epochs, training our network takes roughly 1.7 days with an NVIDIA 1080TI GPU.

\begin{table}
\begin{center}
\scalebox{0.9}{
\begin{tabular}{|l|c|c|c|c|c|c|}
\hline
PSNR& City &Walk &Calendar &Foliage & Avg.\\
\hline\hline
TOFlow &26.77 &29.05 &22.54 &25.32 &25.92\\
TDAN &\textbf{26.97} &\textbf{29.48} &\textbf{22.98} &\textbf{25.50} &\textbf{26.23}\\
\hline
\hline
SSIM & City &Walk &Calendar &Foliage & Avg.\\
\hline\hline
TOFlow &0.740 &0.880 &0.731 &0.710 &0.765\\
TDAN &\textbf{0.756} &\textbf{0.889} &\textbf{0.7556} &\textbf{0.717} &\textbf{0.779}\\
\hline
\end{tabular}
}
\end{center}
\caption{PSNR (dB) and SSIM of the TOFlow~\cite{xue2017video} and our method on Vid4 dataset with an upscale factor 4 on the BI configuration. Note that, the first and last three frames are not used for evaluation since the TOFlow does not predict them; we used the same training dataset as the TOFlow.}
\label{toflow}
\vspace{-5mm}
\end{table}

\begin{table*}
\begin{center}
\scalebox{0.9}{
\begin{tabular}{|l|c|c|c|c|c|c| c| c|c| c|}
\hline
PSNR & Bicubic &VSRnet~\cite{kappeler2016video} &ESPCN~\cite{shi2016real} & VESCPN \cite{caballero2017real} & Liu \etal \cite{liu2017robust} & DBPN~\cite{haris2018deep} &RDN~\cite{zhang2018residual} &RCAN~\cite{zhang2018image} &TDAN \\
\hline\hline
City & 25.13 &25.62 &25.82 &26.17 &\textcolor{blue}{26.50} &25.80 &26.04 &26.06 &\textcolor{red}{\textbf{26.99}}\\
Walk & 26.06 &27.54 &28.05 &28.31 &{28.35} &\textcolor{blue}{28.64} &28.61 &\textcolor{blue}{28.64} &\textcolor{red}{\textbf{29.50}}\\
Calendar &20.54 &21.34 &21.76 &21.98 & 22.13 &22.29 &22.32 &\textcolor{blue}{22.33} &\textcolor{red}{\textbf{22.98}}\\
Foliage& 23.50 &24.41 &24.58 &24.91 & \textcolor{blue}{25.09} & 24.73 &24.76 &24.77 &\textcolor{red}{\textbf{25.51}}\\
Avg. & 23.81 &24.73 &25.06  &25.34 & \textcolor{blue}{25.52} &25.37 &25.43 &25.45 &\textcolor{red}{\textbf{26.24}}\\
\hline
\hline
SSIM & Bicubic &VSRnet~\cite{kappeler2016video} &ESPCN~\cite{shi2016real} & VESCPN \cite{caballero2017real} &Liu \etal \cite{liu2017robust} & DBPN~\cite{haris2018deep} &RDN~\cite{zhang2018residual} &RCAN~\cite{zhang2018image} & TDAN \\
\hline\hline
City & 0.601 &0.654 &0.670 &0.696 &\textcolor{blue}{0.723} & 0.682 &0.688 &0.694 &\textcolor{red}{\textbf{0.757}}\\
Walk & 0.798 &0.844 &0.859 &0.861 &0.863 &0.872 &0.872 &\textcolor{blue}{0.873} &\textcolor{red}{\textbf{0.890}}\\
Calendar & 0.571 &0.644 &0.679 &0.691 &0.707 &0.715 &0.721 &\textcolor{blue}{0.723} &\textcolor{red}{\textbf{0.756}}\\
Foliage& 0.566 &0.645 &0.652 &0.673 &\textcolor{blue}{0.701} &0.661 &0.663 &0.664 &\textcolor{red}{\textbf{0.717}}\\
Avg. & 0.634 &0.697 &0.715 &0.730 &\textcolor{blue}{0.748} &0.732 &0.736 &0.738 &\textcolor{red}{\textbf{0.780}}\\
\hline
\end{tabular}
}
\end{center}
\caption{PSNR (dB) and SSIM of different networks on Vid4 dataset with upscale factor 4 on BI configuration. Note that, the first and last two frames are not used for evaluation as Liu \etal \cite{liu2017robust}. The top-2 results are highlighted with red and blue colors. Compared with recent state-of-the-art SISR and VSR methods on BI configuration, our method can achieve the best performance.}
\label{psnr&ssim}
\vspace{-4mm}
\end{table*}
\subsection{Experimental Comparisons}
\label{exp_comp}

\noindent \textbf{Results with BI Degradation:}
Tables \ref{toflow} and \ref{psnr&ssim} show quantitative comparisons on the BI configuration. Note that our TDAN used the same training dataset as the TOFlow, and other VSR methods did not release their training data or training source code. Therefore we compare with their methods based directly on the provided results. We can see that our TDAN achieves the best performance among all compared state-of-the-art flow-based VSR networks and SISR networks.

\begin{figure*}
	\scriptsize
	\centering
	\begin{tabular}{cc}
        \begin{adjustbox}{valign=t}
			\begin{tabular}{c}
				\includegraphics[width=0.262\textwidth]{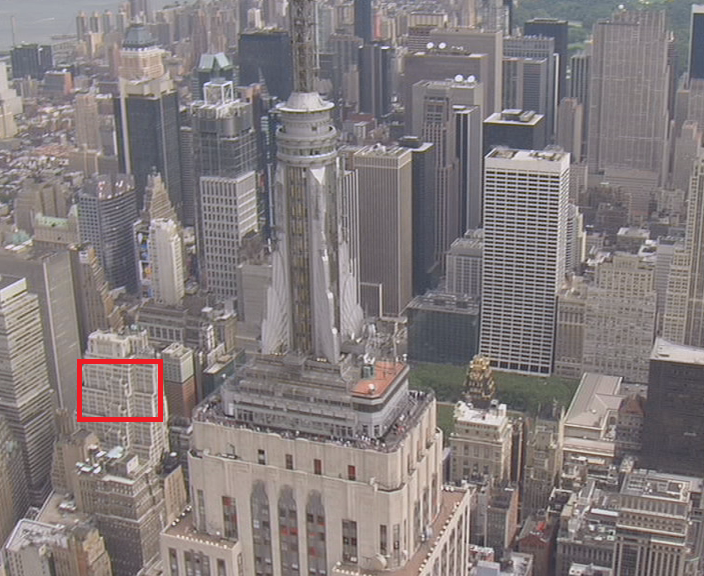}
				\\
				City/BI
			
			\end{tabular}
		\end{adjustbox}
		\hspace{-2.3mm}
		\begin{adjustbox}{valign=t}
			\begin{tabular}{ccccc}
				\includegraphics[width=\widthscalefive \textwidth]{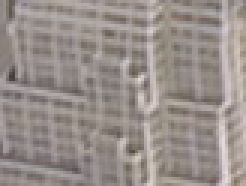} \hspace{-1.5mm} &
				\includegraphics[width=\widthscalefive \textwidth]{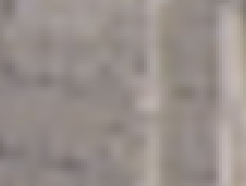} \hspace{-1.5mm} &
				\includegraphics[width=\widthscalefive \textwidth]{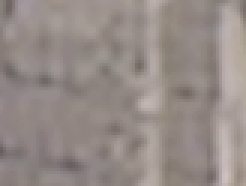} \hspace{-1.5mm} &
				\includegraphics[width=\widthscalefive \textwidth]{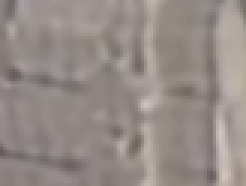}\hspace{-1.5mm} &
				\includegraphics[width=\widthscalefive \textwidth]{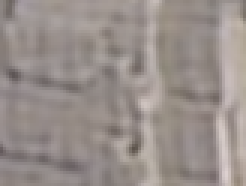}
				\\
				HR \hspace{-1.5mm} &
				Bicubic \hspace{-1.5mm} &
				VSRnet~\cite{kappeler2016video} \hspace{-1.5mm} &
				VESCPN~\cite{caballero2017real}\hspace{-1.5mm} &
				Liu \etal~\cite{liu2017robust}

				\\
				\includegraphics[width=\widthscalefive \textwidth]{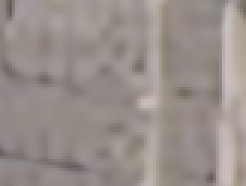} \hspace{-1.5mm} &
				\includegraphics[width=\widthscalefive \textwidth]{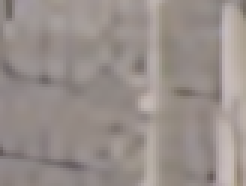} \hspace{-1.5mm} &
				\includegraphics[width=\widthscalefive \textwidth]{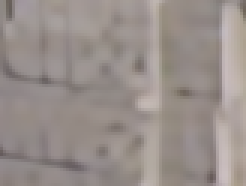} \hspace{-1.5mm} &
				\includegraphics[width=\widthscalefive \textwidth]{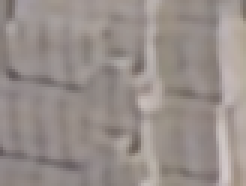}\hspace{-1.5mm} &
				\includegraphics[width=\widthscalefive \textwidth]{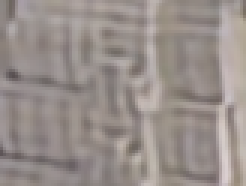}
				\\
				DBPN~\cite{haris2018deep} \hspace{-1.5mm} &
				RDN~\cite{zhang2018residual} \hspace{-1.5mm} &
				RCAN~\cite{zhang2018image} \hspace{-1.5mm} &
				TOFlow~\cite{xue2017video} \hspace{-1.5mm} &
				TDAN
				\\
			\end{tabular}
			\end{adjustbox}

         \\
    \begin{adjustbox}{valign=t}
			\begin{tabular}{c}
				\includegraphics[width=0.262\textwidth]{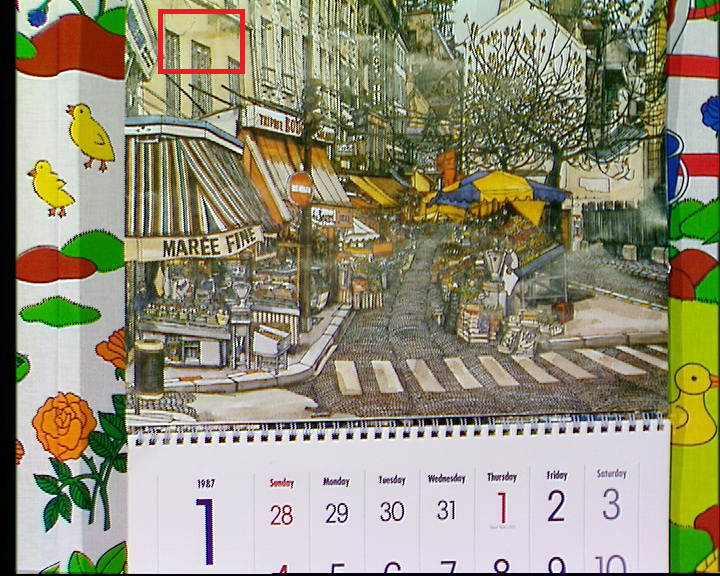}
				\\
				Calendar/BI
			
			\end{tabular}
		\end{adjustbox}
		\hspace{-2.3mm}
		\begin{adjustbox}{valign=t}
			\begin{tabular}{ccccc}
				\includegraphics[width=\widthscalefive \textwidth]{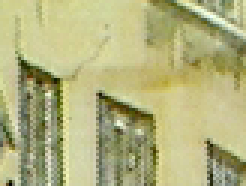} \hspace{-1.5mm} &
				\includegraphics[width=\widthscalefive \textwidth]{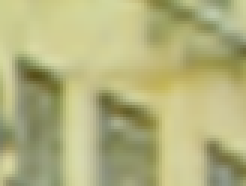} \hspace{-1.5mm} &
				\includegraphics[width=\widthscalefive \textwidth]{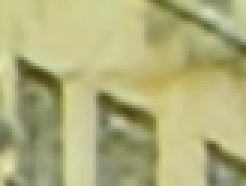} \hspace{-1.5mm} &
				\includegraphics[width=\widthscalefive \textwidth]{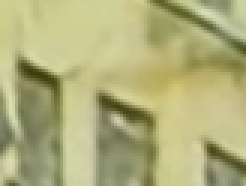}\hspace{-1.5mm} &
				\includegraphics[width=\widthscalefive \textwidth]{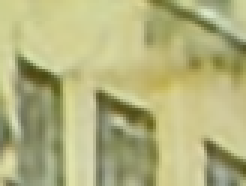}
				\\
				HR \hspace{-1.5mm} &
				Bicubic \hspace{-1.5mm} &
				VSRnet~\cite{kappeler2016video} \hspace{-1.5mm} &
				VESCPN~\cite{caballero2017real}\hspace{-1.5mm} &
				Liu \etal~\cite{liu2017robust}

				\\
				\includegraphics[width=\widthscalefive \textwidth]{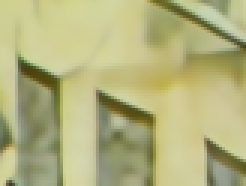} \hspace{-1.5mm} &
				\includegraphics[width=\widthscalefive \textwidth]{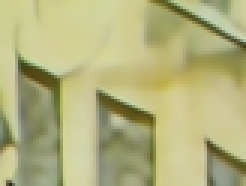} \hspace{-1.5mm} &
				\includegraphics[width=\widthscalefive \textwidth]{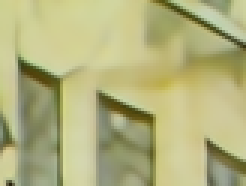} \hspace{-1.5mm} &
				\includegraphics[width=\widthscalefive \textwidth]{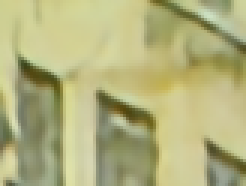}\hspace{-1.5mm} &
				\includegraphics[width=\widthscalefive \textwidth]{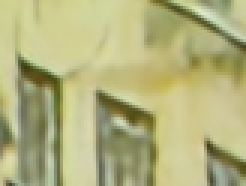}
				\\
				DBPN~\cite{haris2018deep} \hspace{-1.5mm} &
				RDN~\cite{zhang2018residual} \hspace{-1.5mm} &
				RCAN~\cite{zhang2018image} \hspace{-1.5mm} &
				TOFlow~\cite{xue2017video} \hspace{-1.5mm} &
				TDAN
				\\
			\end{tabular}
			\end{adjustbox}

         \\
          \begin{adjustbox}{valign=t}
			\begin{tabular}{c}
				\includegraphics[width=0.262\textwidth]{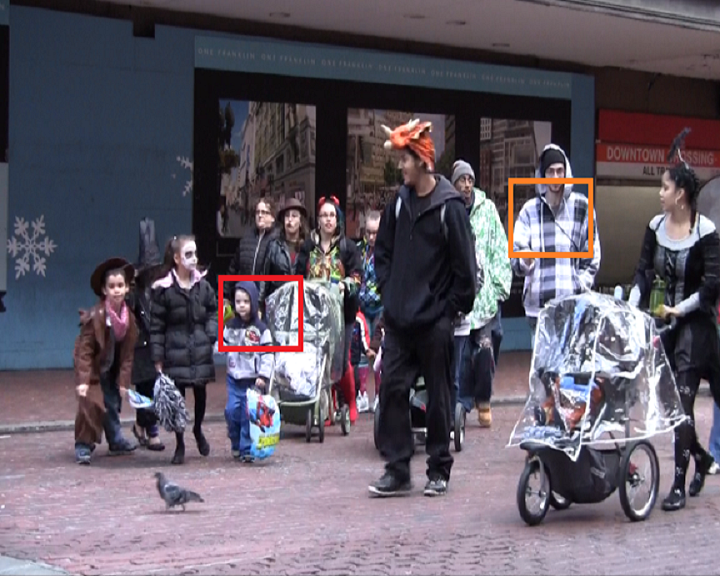}
				\\
				Walk/BD
			
			\end{tabular}
		\end{adjustbox}
		\hspace{-2.3mm}
		\begin{adjustbox}{valign=t}
			\begin{tabular}{ccccc}
				\includegraphics[width=\widthscalefive \textwidth]{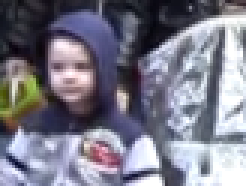} \hspace{-1.5mm} &
				\includegraphics[width=\widthscalefive \textwidth]{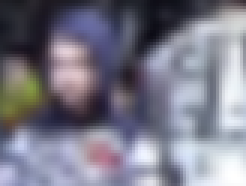} \hspace{-1.5mm} &
				\includegraphics[width=\widthscalefive \textwidth]{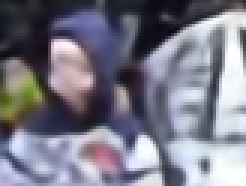} \hspace{-1.5mm} &
				\includegraphics[width=\widthscalefive \textwidth]{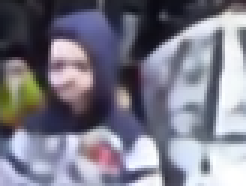}\hspace{-1.5mm} &
				\includegraphics[width=\widthscalefive \textwidth]{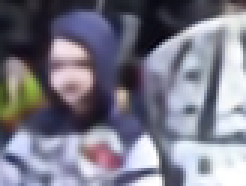}
				\\
				HR \hspace{-1.5mm} &
				Bicubic \hspace{-1.5mm} &
				SPMC~\cite{tao2017detail} \hspace{-1.5mm} &
				DUF~\cite{jo2018deep}\hspace{-1.5mm} &
				TDAN

				\\
				\includegraphics[width=\widthscalefive \textwidth]{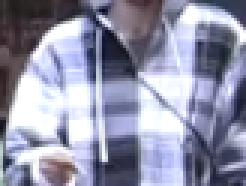} \hspace{-1.5mm} &
				\includegraphics[width=\widthscalefive \textwidth]{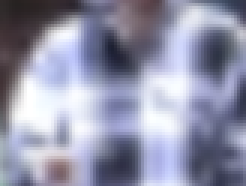} \hspace{-1.5mm} &
				\includegraphics[width=\widthscalefive \textwidth]{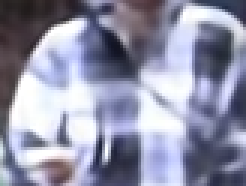} \hspace{-1.5mm} &
				\includegraphics[width=\widthscalefive \textwidth]{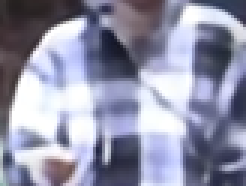}\hspace{-1.5mm} &
				\includegraphics[width=\widthscalefive \textwidth]{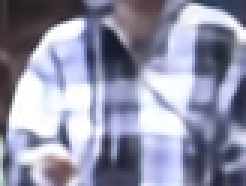}
				\\
				HR \hspace{-1.5mm} &
				Bicubic \hspace{-1.5mm} &
				SPMC~\cite{tao2017detail} \hspace{-1.5mm} &
				DUF~\cite{jo2018deep}\hspace{-1.5mm} &
				TDAN
				\\
			\end{tabular}
			\end{adjustbox}

         \\
          \begin{adjustbox}{valign=t}
			\begin{tabular}{c}
				\includegraphics[width=0.262\textwidth]{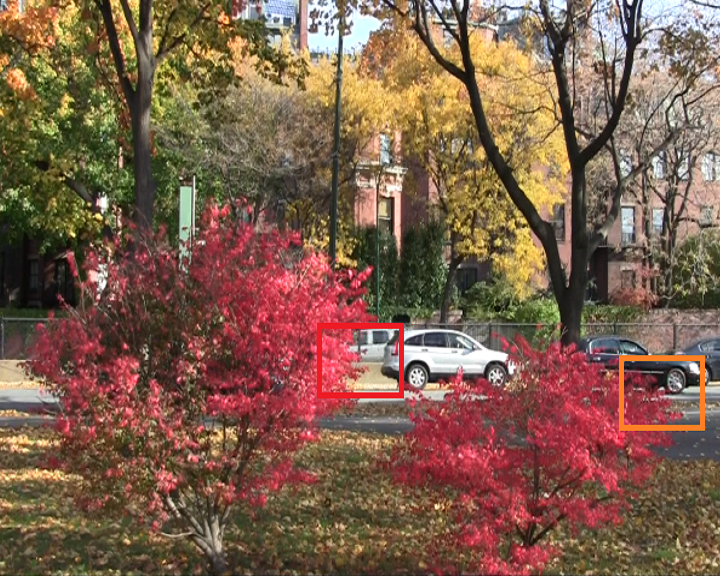}
				\\
				Foliage/BD
			
			\end{tabular}
		\end{adjustbox}
		\hspace{-2.3mm}
		\begin{adjustbox}{valign=t}
			\begin{tabular}{ccccc}
				\includegraphics[width=\widthscalefive \textwidth]{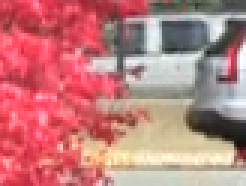} \hspace{-1.5mm} &
				\includegraphics[width=\widthscalefive \textwidth]{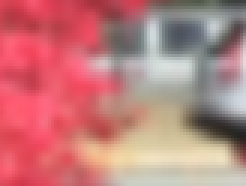} \hspace{-1.5mm} &
				\includegraphics[width=\widthscalefive \textwidth]{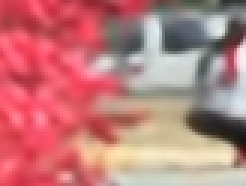} \hspace{-1.5mm} &
				\includegraphics[width=\widthscalefive \textwidth]{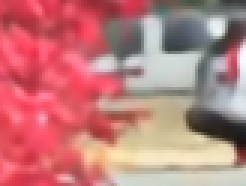}\hspace{-1.5mm} &
				\includegraphics[width=\widthscalefive \textwidth]{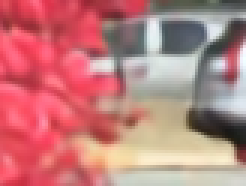}
				\\
				HR \hspace{-1.5mm} &
				Bicubic \hspace{-1.5mm} &
				SPMC~\cite{tao2017detail} \hspace{-1.5mm} &
				DUF~\cite{jo2018deep}\hspace{-1.5mm} &
				TDAN
				\\
				\includegraphics[width=\widthscalefive \textwidth]{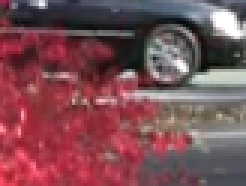} \hspace{-1.5mm} &
				\includegraphics[width=\widthscalefive \textwidth]{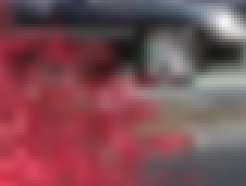} \hspace{-1.5mm} &
				\includegraphics[width=\widthscalefive \textwidth]{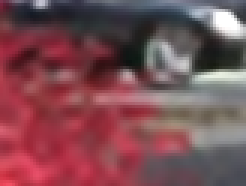} \hspace{-1.5mm} &
				\includegraphics[width=\widthscalefive \textwidth]{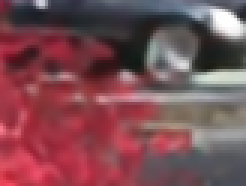}\hspace{-1.5mm} &
				\includegraphics[width=\widthscalefive \textwidth]{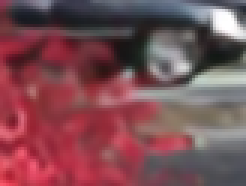}
				\\
				HR \hspace{-1.5mm} &
				Bicubic \hspace{-1.5mm} &
				SPMC~\cite{tao2017detail} \hspace{-1.5mm} &
				DUF~\cite{jo2018deep}\hspace{-1.5mm} &
				TDAN
				\\
			\end{tabular}
			\end{adjustbox}

         \\
            \vspace{-1mm}

	\end{tabular}
	\caption{Visual comparisons for $4\times$ VSR on the BI and BD configurations. We observe that the proposed TDAN restores better image structures and details than other state-of-the-art VSR networks, which demonstrates the strong capability of the TDAN in temporal alignment leveraging informative pixels from LR supporting frames.}
	\label{fig:result_x4}
\end{figure*}

Visual results of \textit{City} and \textit{Calendar} for $4\times$ VSR on the BI configuration are illustrated in Fig.~\ref{fig:result_x4}.  We can find that the SISR networks without using the supporting frames: DPBN, RDN, and RCAN, fail to restore missing image details such as the building structures in \textit{City} and the window in \textit{Calendar}. With motion compensation, the VESCPN~\cite{caballero2017real}, Liu \etal~\cite{liu2017robust}, and TOFlow~\cite{xue2017video} can compensate missing details from the supporting frames. Our TDAN recovers more fine image details than others, which demonstrates the effectiveness of the proposed one-stage temporal alignment framework.

\noindent \textbf{Results with BD Degradation:}
Table \ref{BD} shows quantitative comparisons on the BD configuration. Our method outperforms the flow-based networks: SPMC and recent FRVSR method but is worse than the dynamic filtering-based DUF on SSIM; the latter implicitly handles the motion by learning to output dynamic upsampling filters and utilizes a large HR training dataset with an additional temporal augmentation. The DUF does not well restore borders of frames; thus the PSNR value of DUF is lower than the TDAN. In addition, the DUF and the FRVSR take 7 and 10 frames as inputs respectively, and our TDAN with 5 frames as inputs exploits less sequence information.

Visual results of \textit{Walk} and \textit{Foliage} for $4\times$ VSR on the BD configuration are illustrated in Fig.~\ref{fig:result_x4}. In comparison with SPMC~\cite{tao2017detail} and DUF~\cite{jo2018deep}, benefiting from the strong temporal alignment network, the proposed TDAN is more effective in exploiting information in supporting frames, and therefore it is more capable of restoring image structures, for example, the baby face and the stripe on the cloth in \textit{Walk}, and the white and black cars in \textit{Foliage}.

\noindent \textbf{Model Sizes:} Table \ref{size} shows parameter numbers of several networks with the leading VSR performance. We can see that the state-of-the-art SISR networks: RDN, RCAN, and TOFlow, have larger model sizes than the TDAN, and the proposed TDAN has comparable parameter numbers with the FRVSR and DUF. Even with such a light-weight model, the proposed TDAN still achieves promising VSR performance, which further validates the effectiveness of the proposed one-stage temporal alignment framework.

\begin{table}
\begin{center}
\scalebox{0.8}{
\begin{tabular}{|l|c|c|c|c|c|c|}
\hline
Vid4& Bicubic &SPMC~\cite{tao2017detail} &FSRVSR~\cite{sajjadi2018frame} &DUF~\cite{jo2018deep} & TDAN\\
\hline\hline
PSNR &23.47 &26.05 &26.17 &26.21 &\textbf{26.58}\\
SSIM &0.616 &0.776 &0.798 &\textbf{0.814} &0.801\\
\hline
\end{tabular}
}
\end{center}
\caption{PSNR (dB) and SSIM of different networks with upscale factor 4 on the BD configuration. Values of the FSRVSR are from the original paper.}
\label{BD}
\end{table}
\begin{table}
\begin{center}
\scalebox{0.8}{
\begin{tabular}{|l|c|c|c|c|c|c|}
\hline
Methods& RDN &RCAN &TOFlow &FRVSR &DUF & TDAN\\
\hline\hline
Param./M &22.30 &15.50 &6.20 &2.00 &1.90 &1.97\\
\hline
\end{tabular}
}
\end{center}
\caption{Parameter numbers ($\times10^6$) of several networks with leading VSR performance.}
\label{size}
\vspace{-5mm}
\end{table}
\begin{figure}[h]
\centering
\includegraphics[width=0.5\textwidth]{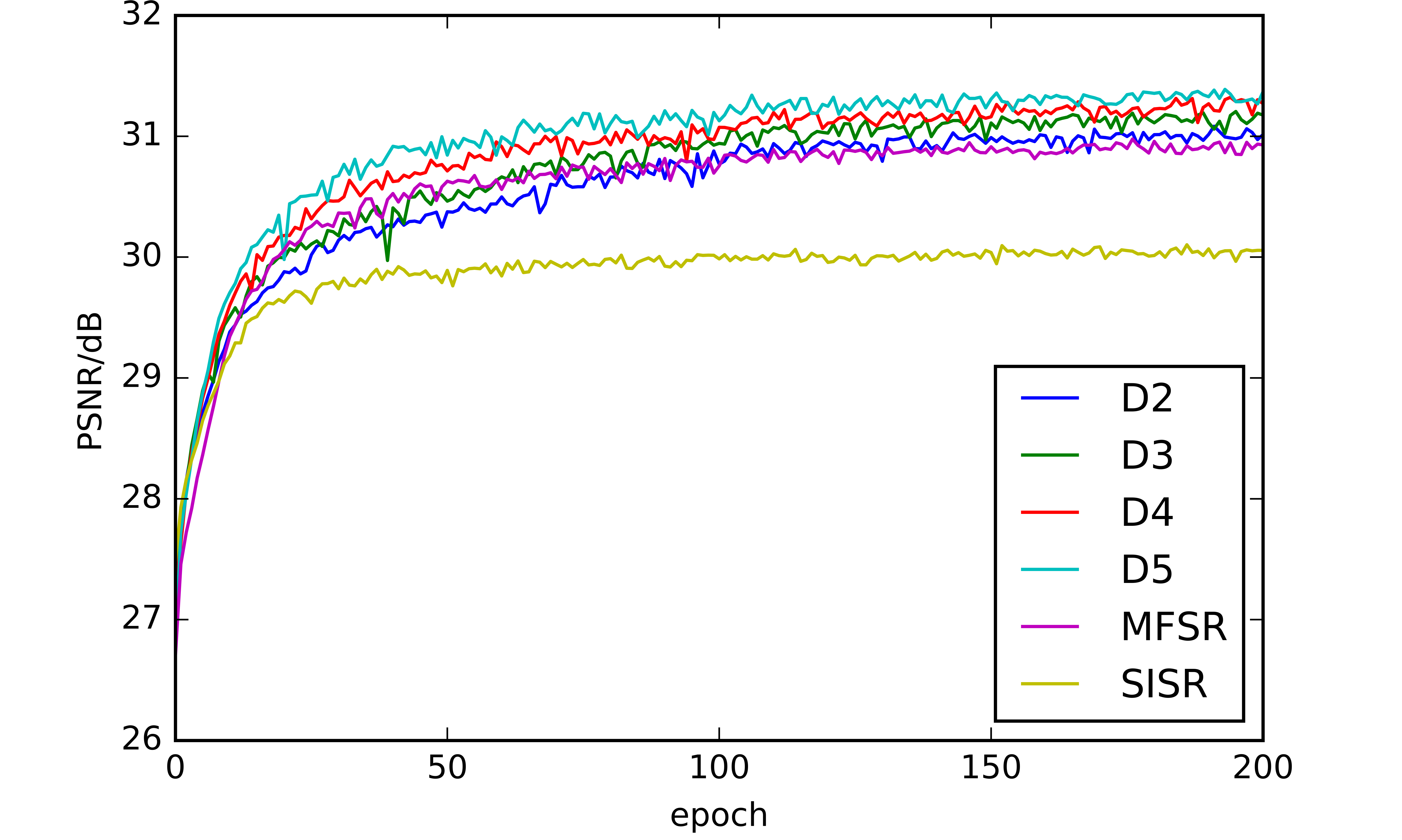}
\caption{Convergence analysis of TDAN with different numbers of the deformable convolutional layers and two baselines: SISR and MFSR. The curves show PSNR values of different models on the validation set in 200 epochs.}
\label{curve}
\end{figure}

\subsection{Ablation Study}
\label{ablation}
To further investigate the proposed TDAN, we compare it with two baselines: SISR and MFSR models, that trained on the Vimeo Super-Resolution dataset. The SISR model only uses the reference frame as the input, and the MFSR directly concatenates the supporting and reference frames as inputs without temporal alignments. The SISR and MFSR networks have similar network structures with the proposed SR reconstruction network in Sec.~\ref{recon}. In addition, we compare our TDAN models with different numbers: 2, 3, 4, and 5 of deformable convolutional layers, and we denote these networks as D2, D3, D4, and D5, respectively.

Figure \ref{curve} illustrates convergence curves of the SISR, MFSR, D2, D3, D4, and D5 networks. We can see that the MFSR outperforms the SISR; D2, D3, D4, and D5 are better than the MFSR; more deformable layers in the proposed TDAN the better VSR performance. These observations demonstrate that exploiting supporting frames even without temporal alignment can improve VSR performance; the proposed temporal alignment network is effective in utilizing information from supporting frames; more deformable layers can enhance the capability of TDAN. For setting TDAN with comparable model size as FRVSR and DUF, we used D4 in our experiments. From qualitative and quantitative comparisons in Sec.~\ref{exp_comp}, we find that even D4 has achieved state-of-the-art VSR performance.

\begin{figure}
\includegraphics[width=0.48\textwidth]{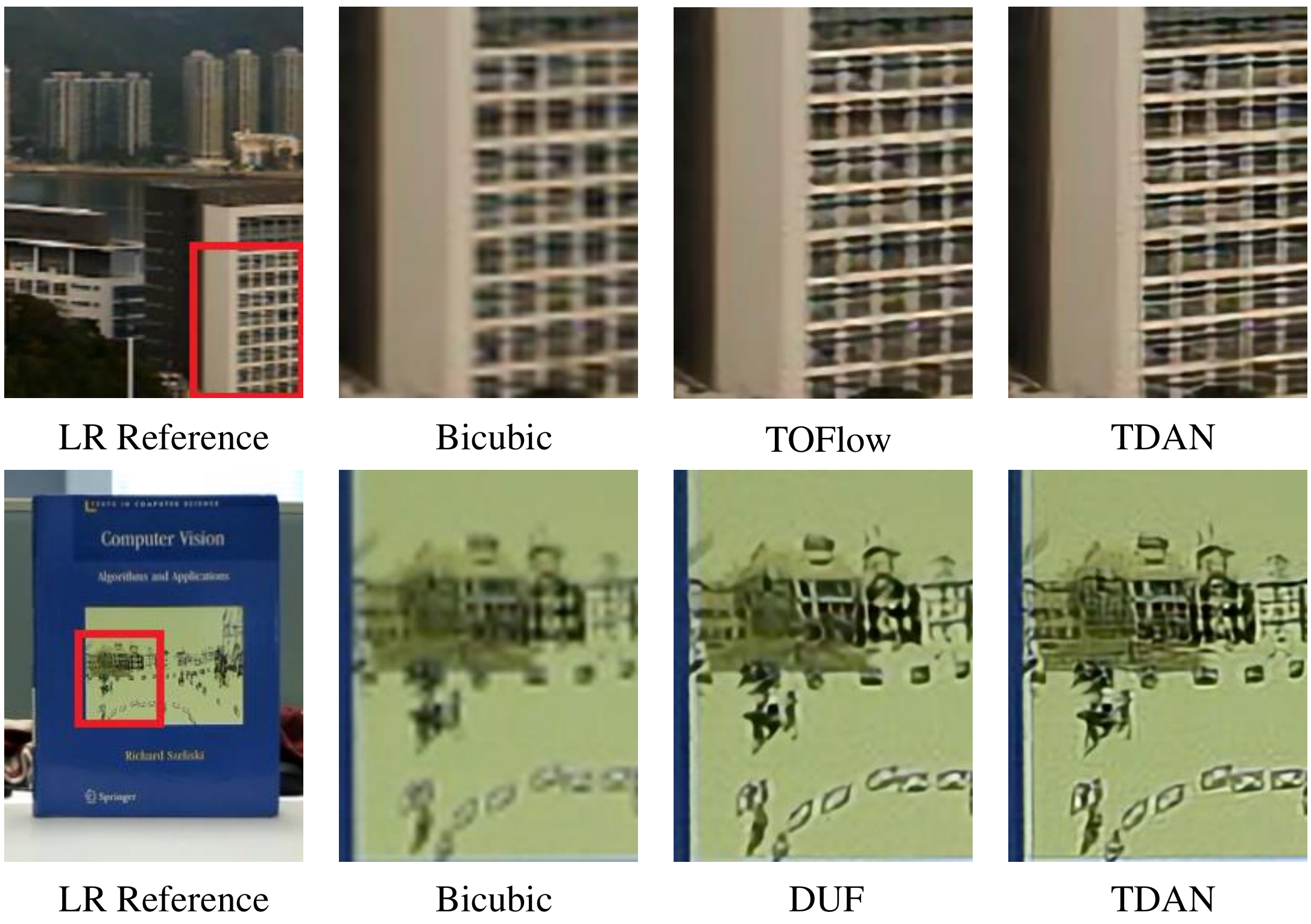}
\caption{Visual results on real-world video sequences with scaling factor 4. The two rows show VSR results for videos \textit{bldg} and \textit{CV Book} from~\cite{Liao_2015_ICCV}, respectively.}
	\label{fig:real}
\end{figure}
\begin{figure}
	\scriptsize
	\centering

		\begin{adjustbox}{valign=t}
			\begin{tabular}{cccc}
				\includegraphics[width=0.11\textwidth]{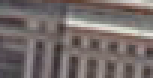} \hspace{-3mm} &
				\includegraphics[width=0.11\textwidth]{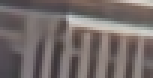}\hspace{-3mm} &
				\includegraphics[width=0.11\textwidth]{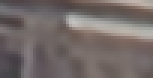}\hspace{-3mm} &
				\includegraphics[width=0.11\textwidth]{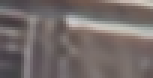}
				\\
				HR \hspace{-3mm} &
				RCAN \hspace{-3mm} &
				TOFlow \hspace{-3mm} &
				TDAN
				\\
			\end{tabular}
			\end{adjustbox}
	\caption{A failure case of the TDAN. The very deep SISR network: RCAN trained on the DIV2K can accurately recover the structures of the shown image region in the \textit{city} video frame, but TOFlow and TDAN failed.}
	\label{failure}
\end{figure}

\subsection{Real-World Examples}
\label{real}
To further validate the robustness of the proposed TDAN, we also conduct experiments on two real-world video sequences: \textit{bldg} and \textit{CV Book} from~\cite{Liao_2015_ICCV}. For the two sequences, the HR video frames are not available, and the degradation methods are unknown. We compare the proposed TDAN with TOFlow on the BI configuration and DUF on the BD configuration. As illustrated in Fig.~\ref{fig:real}, we find that the proposed TDAN can produce sharper edges (see \textit{bldg} results) and more image details (see \textit{CV Book} results) than the compared state-of-the-art VSR networks. The comparison results show that our TDAN can robustly handle even unknown degradation, which further demonstrates the superiority of the proposed framework.

\section{Limitation and Failure Exploration}
\label{lim_fail}

As discussed in Sec.~\ref{exp_comp}, the resolution of HR video frames in Vimeo Super-Resolution dataset is only $448\times256$. It is hard to train a very deep network on the dataset for recovering even finer image structures and details. One failure case of the TDAN is shown in Fig.~\ref{failure}. We can see that the TDAN fails to recover the structures in the building, but the very deep SISR network: RCAN trained on the DIV2K dataset can accurately reconstruct them, which demonstrates that the LR reference frame can provide enough cues for restoring the structures without requiring additional information from the LR supporting frames. Therefore, it is worth to build a publicly available large VSR dataset with HR (\eg 1080P, 2K, even 4K) video frames for training very deep VSR architectures.

In Sec.~\ref{recon}, we adopt a naive and simple method to perform temporal fusion, which directly concatenates a LR reference frame and the corresponding LR aligned frames, and then uses a convolutional layer to obtain temporally fused features. A more advanced fusion network may further improve the VSR performance of the TDAN.

In our initial design, we directly took the output feature map $F_i^{LR'}$ (see Eq. \ref{theta_eq}) of the deformable alignment module and the feature map $F_t^{LR}$ of the LR reference frame as inputs to a SR reconstruction network without predicting the aligned frames and the loss term $\mathcal{L}_{align}$. With this design, it will become a unified VSR framework without explicitly intermediate temporal alignment. However, the performance of the framework is even worse than the baseline model: MFSR. Unlike high-level problems in \cite{Dai_2017_ICCV,Bertasius_2018_ECCV}, the deformable convolution taking features from pre-trained strong backbones as inputs is much easier to learn to capture motions of objects. Therefore, we think that the loss $\mathcal{L}_{align}$ is important for this end-to-end VSR network.

In the TDAN, we use the LR reference frame as the label to define the $\mathcal{L}_{align}$. However, the LR reference frame is not exactly same as real aligned LR frames, which will make the label be noisy. Robust algorithms like~\cite{natarajan2013learning} for learning under label noise can be considered to further improve the $\mathcal{L}_{align}$.

\section{Conclusion}
In this paper, we propose a one-stage temporal alignment network: TDAN for video super-resolution. Unlike previous optical flow-based methods, which splits the temporal alignment problem into two sub-problems: motion estimation and motion compensation, the TDAN implicitly captures motion cues via a deformable sampling module at the feature level and directly predicts aligned LR video frames from sampled features without image-wise wrapping operations. In addition, the TDAN is capable of exploring image contextual information. With the advanced one-stage temporal alignment design and the strong exploration capability, the proposed TDAN-based VSR framework outperforms the compared flow-based state-of-the-art VSR networks. In the future, we would like to adopt the proposed TDAN to solve other video restoration tasks, such as video denoising, video deblurring, and video frame interpolation.

\noindent \textbf{Acknowledgement}
This work was supported in part by NSF IIS-1813709, IIS-1741472, and CHE-1764415. Any opinions, findings, and conclusions or recommendations expressed in this material are those of the authors and do not necessarily reflect the views of the NSF.
{\small
\bibliographystyle{ieee}
\bibliography{egbib}
}
\end{document}